\pdfoutput=1

\documentclass[11pt]{article}

\usepackage[]{acl}

\usepackage{times}
\usepackage{latexsym}

\usepackage[T1]{fontenc}

\usepackage[utf8]{inputenc}

\usepackage{microtype}

\usepackage{amsmath}
\usepackage{amsfonts}
\usepackage{graphicx}
\usepackage{algorithm}
\usepackage{algorithmic}
\usepackage{paralist}
\usepackage{subfigure}
\usepackage{stfloats}  

\title{Alleviating the Inequality of Attention Heads \\ for Neural Machine Translation}

\author{
    Zewei Sun\textsuperscript{\rm 1,\thanks{* Work was done while at NJU}}, 
    Shujian Huang\textsuperscript{\rm 2,3},
    Xin-Yu Dai\textsuperscript{\rm 2}, 
    Jiajun Chen\textsuperscript{\rm 2} \\
    \textsuperscript{\rm 1} ByteDance AI Lab \\
    \textsuperscript{\rm 2} State Key Laboratory for Novel Software Technology, Nanjing University \\
    \textsuperscript{\rm 3} Peng Cheng Laboratory, China \\
    \texttt{sunzewei.v@bytedance.com}, 
    \texttt{\{huangsj,daixinyu,chenjj\}@nju.edu.cn}
}

\begin{document}
\maketitle
\begin{abstract}
Recent studies show that the attention heads in Transformer are not equal~\cite{Voita2019AnalyzingMS,Michel2019AreSH}.
We relate this phenomenon to the imbalance training of multi-head attention and the model dependence on specific heads. To tackle this problem, we propose a simple masking method: \textit{HeadMask}, in two specific ways. Experiments show that translation improvements are achieved on multiple language pairs. Subsequent empirical analyses also support our assumption and confirm the effectiveness of the method.
\end{abstract}

\section{Introduction}

Recently, more and more novel network structures of neural machine translation(NMT) have been proposed \cite{Bahdanau2015NeuralMT,Barone2017DeepAF,Gehring2017ConvolutionalST,Vaswani2017AttentionIA}, among which Transformer \cite{Vaswani2017AttentionIA} achieves the best results. One important difference between Transformer and other translation models is its multi-head attention mechanism. 

Some interesting phenomena of the attention heads are discovered recently. \newcite{Voita2019AnalyzingMS} find that only a small subset of heads appear to be important for the translation task and vast majority of heads can be removed without seriously affecting performance. \newcite{Michel2019AreSH} also find that several heads can be removed from trained transformer models without statistically significant degradation in
test performance. It turns out that not all heads are equally important.

We speculate that this can be attributed to the imbalanced training of multi-head attention, as some heads are not trained adequately and contribute little to the model. However, this can be turned into the bottleneck for the whole model. For an analogy, if a soccer player gets used to using the right foot and spares more training opportunities for it, it will be stronger and stronger. As a result, the right foot is further relied on, while the left foot receives less training and gradually turns into the limitation.

In this paper, we firstly empirically confirm the inequality in multi-head attention. Then a new training method with two variants 
is proposed to avoid the bottleneck and improve the translation performance. Further analyses are also made to verify the assumption.


\section{Head Inequality}

Following~\newcite{Michel2019AreSH}, we define the importance of an attention head $h$ as 




\begin{equation}
    I_h = \mathbb{E}_{x \sim X} \left| \frac{\partial \mathcal{L}(x)}{\partial \xi_h} \right|
    \label{rbst_eq_importance}
\end{equation}


where $\mathcal{L}(x)$ is the loss on sample x and $\xi$ is the head mask variable with values in \{0, 1\}. Intuitively, if $head_h$ is important, switching $\xi_{h}$ will have a significant effect on the loss. Applying the chain rule yields the final expression for $I_h$:

\begin{equation}
    I_h = \mathbb{E}_{x \sim X} \left| \text{Att}_{h}(x)^T \frac{\partial \mathcal{L}(x)}{\partial \text{Att}_{h}(x)} \right|
\end{equation}

This is equivalent to the Taylor expansion method from~\newcite{molchanov2017pruning}.
In Transformer base~\cite{Vaswani2017AttentionIA}, there are 3 types of attention (encoder self attention, decoder self attention, encoder-decoder attention) with 6 layers per type and 8 heads per layer. Therefore, it amounts to 144 heads. 
We divide them into 8 groups with 18 heads (12.5\%) each group according to their importance $I_h$, among which, 1-18 are the most important and so on.

\begin{figure}[h]
    \centering
    \includegraphics[scale = 0.4]{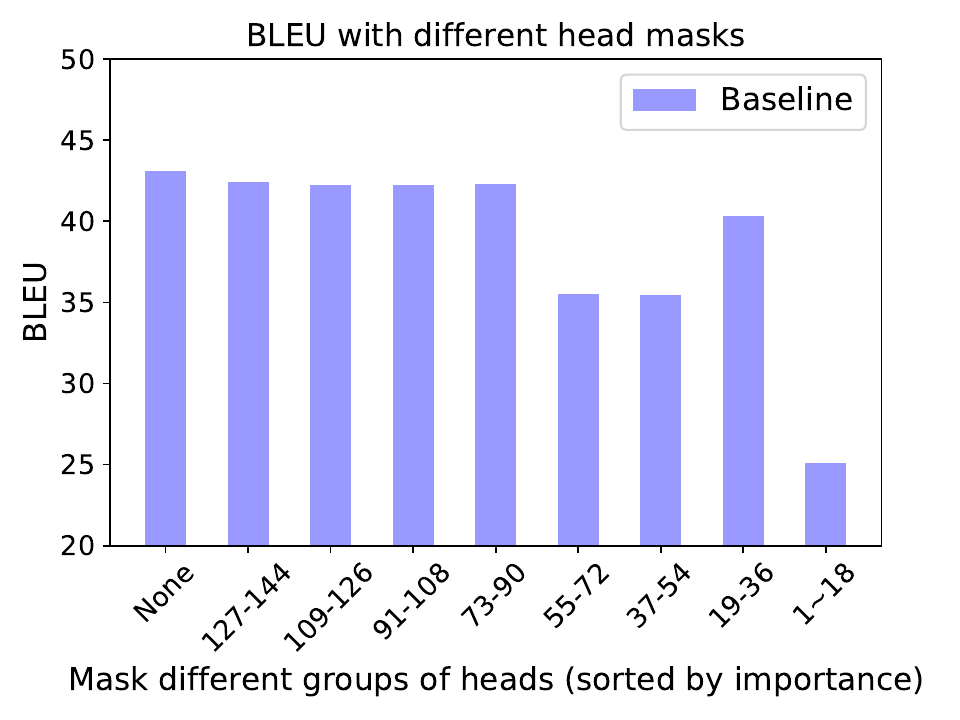}
    \caption{Mask the heads in the same group. Important ones matter much more than unimportant ones.}
    \label{fig_maskhead}
\end{figure}

We then mask different groups of the heads. As is shown in Figure~\ref{fig_maskhead}, masking a group of unimportant heads has little effect on the translation quality while masking important heads leads to a significant drop of performance. Surprisingly, almost half of the heads are not important, as it makes almost no difference whether they are masked or not.


\begin{figure}[h]
    \centering
    \includegraphics[scale = 0.4]{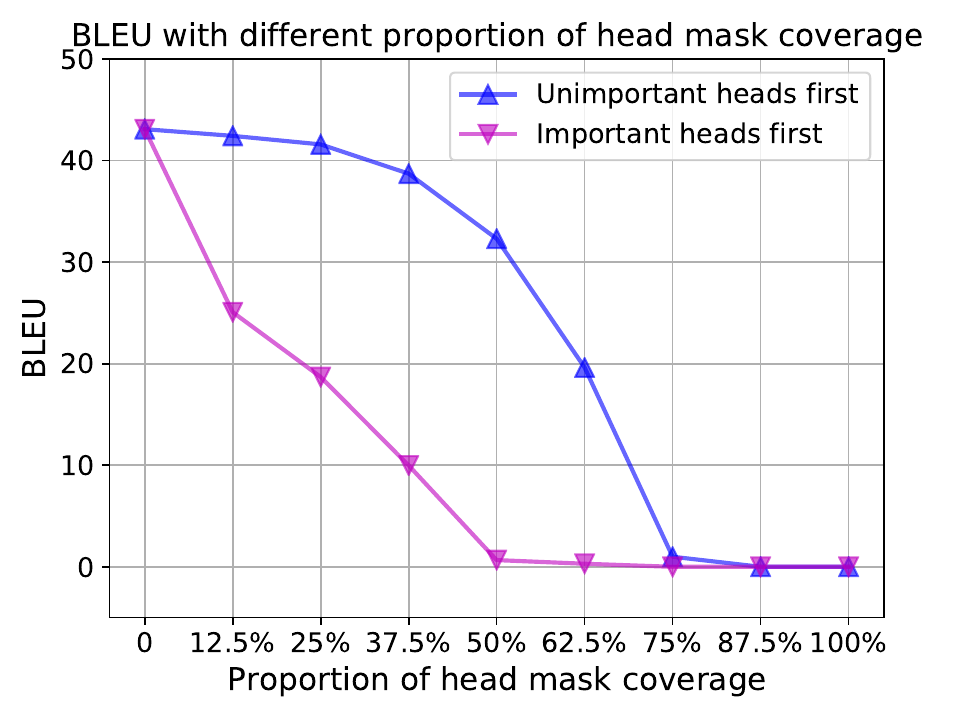}
    \caption{Mask all heads in the ascending order and descending order. The drop curves differ greatly.} 
    \label{fig_reduce}
\end{figure}

We also gradually masking more heads group by group in the ascending order and descending order, respectively. As is shown in Figure~\ref{fig_reduce}, the line starting with unimportant heads drops much slower than the one starting with important ones. It fully illustrates the inequality of different heads.

Figure~\ref{fig_maskhead} and Figure~\ref{fig_reduce} further demonstrates the inequality of the importance of attention heads. 
A simple assumption for explanation is that some heads coincidentally get more updating opportunities in the early stage, which makes the model learning to depend on them gradually. 
As a result, the model increasingly draws a strong connection with these specific heads while this local dependence prevents the rest attention heads from adequate training and restricts the overall capacity. 

\begin{figure*}[b!]\footnotesize
    \centering
    \subfigure[Baseline]{
        \begin{minipage}{.32\textwidth}
            \centering
            \includegraphics[scale = 0.285]{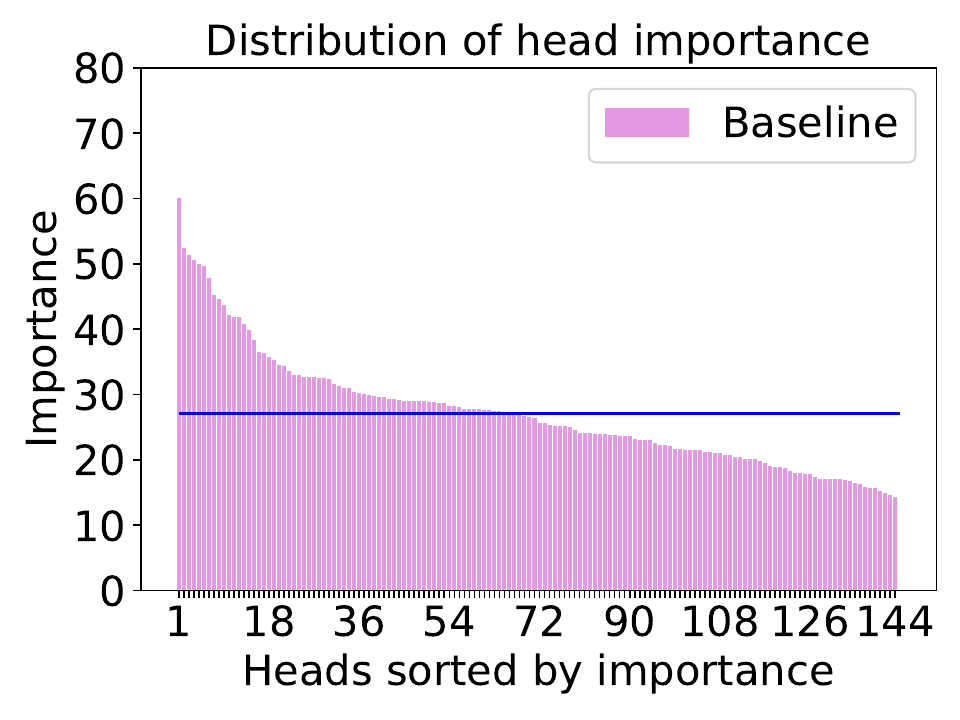}
        \end{minipage}
    }
    \subfigure[Random-18]{
        \begin{minipage}{.32\textwidth}
            \centering
            \includegraphics[scale = 0.285]{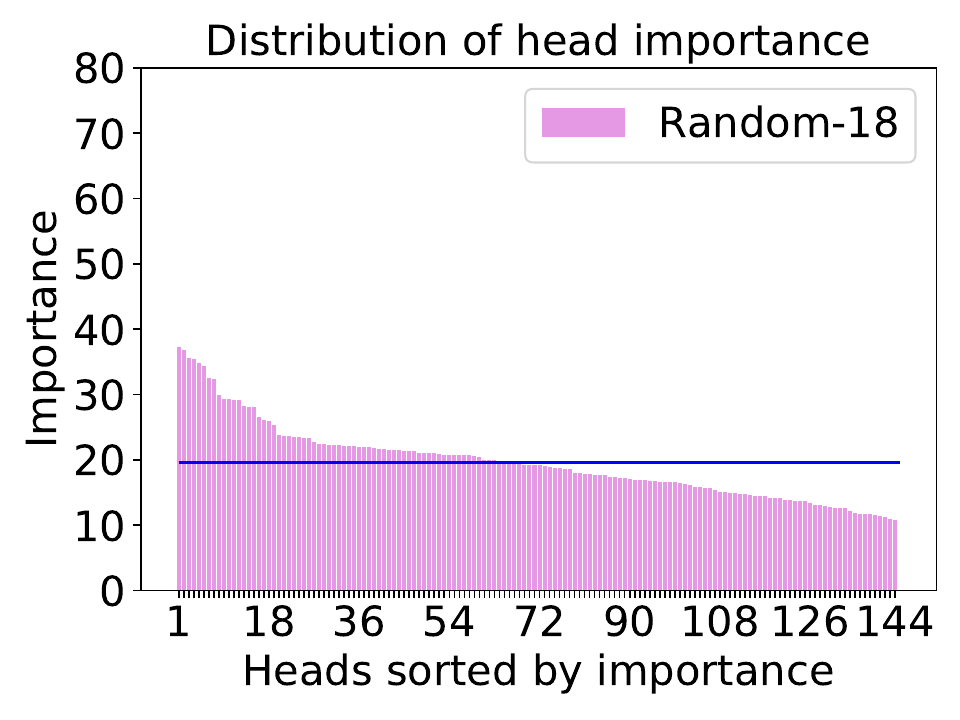}
        \end{minipage}
    }
    \subfigure[Impt-18]{
        \begin{minipage}{.32\textwidth}
            \centering
            \includegraphics[scale = 0.285]{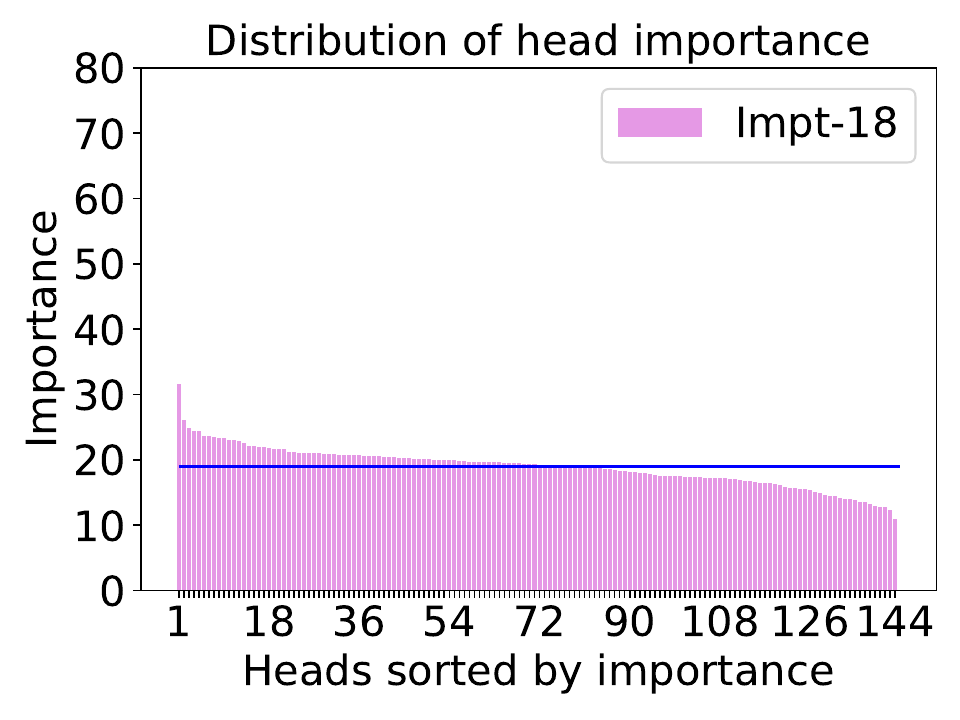}
        \end{minipage}
    }
    \caption{Distribution of importance of attention heads. Our methods make the whole distribution much flatter.}
    \label{fig_maskhead_dis}
\end{figure*}

\section{HeadMask}

Since the problem refers to the unfair training of attention heads, it is natural for us to explicitly balance the training chances. We propose a simple method: \textit{HeadMask}, which masks certain heads during training in two specific ways.

\subsection{Mask Randomly}

The first one is randomly picking heads and masking them in each batch. It ensures every head gets relatively equal opportunities of training and avoid partial dependence, as is shown in Algorithm~\ref{alg_random}. For the soccer analogy, it is like training the feet randomly, making both receive the same amount of practice.

\begin{algorithm}[h]\footnotesize
    \caption{HeadMask: Mask Randomly}
    \label{alg_random}
    \begin{flushleft}
        \hspace*{0.02in} {\bf Input:} $q, k, v$ for attention, number of masks $n$\\
        \hspace*{0.02in} {\bf Output:} masked context
    \end{flushleft}
    \begin{algorithmic}[1]
        \FOR {batch in datasets}
            \STATE heads = random.sample(all\_heads, n)
            \FOR {head in heads}
                \STATE $\xi_{head} = 0$
            \ENDFOR
            \STATE context = attn($\xi$)
        \ENDFOR
    \end{algorithmic}
\end{algorithm}

\subsection{Mask Important Ones}

The second one is masking the most important heads. By forcing the model neglects important heads, we hope more training chances are assigned to weaker heads. For the soccer analogy, it means training the left foot more if the right foot dominates. And once reversed, train contrarily. Its main idea is about suppressing addicted training. Specifically, the network firstly proceeds feed-forward calculation and back propagation without updating parameters to yield the importance of heads. And after picking the most important heads by sorting, mask them. During training, we only use the rest part of networks to reach the final loss and update parameters, as is shown in algorithm~\ref{alg_importance}.

\begin{algorithm}[h]\footnotesize
    \caption{HeadMask: Mask Important Ones}
    \label{alg_importance}
    \begin{flushleft}
        \hspace*{0.02in} {\bf Input:} $q, k, v$ for attention, number of masks $n$\\
        \hspace*{0.02in} {\bf Output:} masked context
    \end{flushleft}
    \begin{algorithmic}[1]
        \FOR {batch in datasets}
            \STATE calculate $\mathcal{L}$ by feed-forward
            \STATE back propagation without updating params
            \STATE calculate importance of all heads $I$
            \STATE heads = $\text{argmax}_n$($I$)
            \FOR {head in heads}
                \STATE $\xi_{head} = 0$
            \ENDFOR
            \STATE context = attn($\xi$)
            \STATE calculate $\mathcal{L}$ by feed-forward
            \STATE back propagation and update params
        \ENDFOR
  \end{algorithmic}
\end{algorithm}

\section{Experiments}

\subsection{Datasets and Systems}

We conduct experiments on four datasets, including three low-resource ones (less than 1 million). We use BPE~\cite{sennrich2016bpe} for Zh-En~\cite{Zheng2018LearningTD} and Ro-En, adopt the preprocessed versions from~\newcite{luong2015stanford} as well as the settings of~\newcite{huang2017neural} for Vi-En, and follow the joint-BPE settings of~\newcite{Sennrich2017TheUO} for Tr-EN.
More information is in Table~\ref{tab_dataset}.



\begin{table}[h]\footnotesize
    \centering
    \begin{tabular}{lccc}
        \hline
        Datasets & Scale & Dev & Test \\ \hline
        NIST Zh-En & 1.34M & MT03 & MT04/05/06 \\ \hline
        WMT16 Ro-En & 608K & newstest2015 & newstest2016  \\ \hline
        IWSLT15 Vi-En & 133K & tst2012 & tst2013 \\ \hline
        WMT17 Tr-En & 207K & newstest2016 & newstest2017 \\ \hline
    \end{tabular}
    \caption{The information of our datasets}
    \label{tab_dataset}
\end{table}


We follow Transformer base setting~\cite{Vaswani2017AttentionIA,sun2022rethinking}.
Parameters are optimized by Adam~\cite{Kingma2015AdamAM}, with $\beta_1 = 0.9$, $\beta_2 = 0.98$, and $\epsilon = 10^{-9}$. The learning rate is scheduled according to \citet{Vaswani2017AttentionIA}, with $warmup\_steps = 4000$. Label smoothing~\cite{Szegedy2016RethinkingTI} of value=0.1 and dropout~\cite{Srivastava2014DropoutAS} of value=0.1 are also adopted.

\begin{figure*}[h]
    \centering
    \subfigure[Baseline]{
        \begin{minipage}{.31\textwidth}
            \centering
            \includegraphics[scale = 0.3]{figures/topdown_baseline.pdf}
        \end{minipage}
    }
    \subfigure[Random-18]{
        \begin{minipage}{.31\textwidth}
            \centering
            \includegraphics[scale = 0.3]{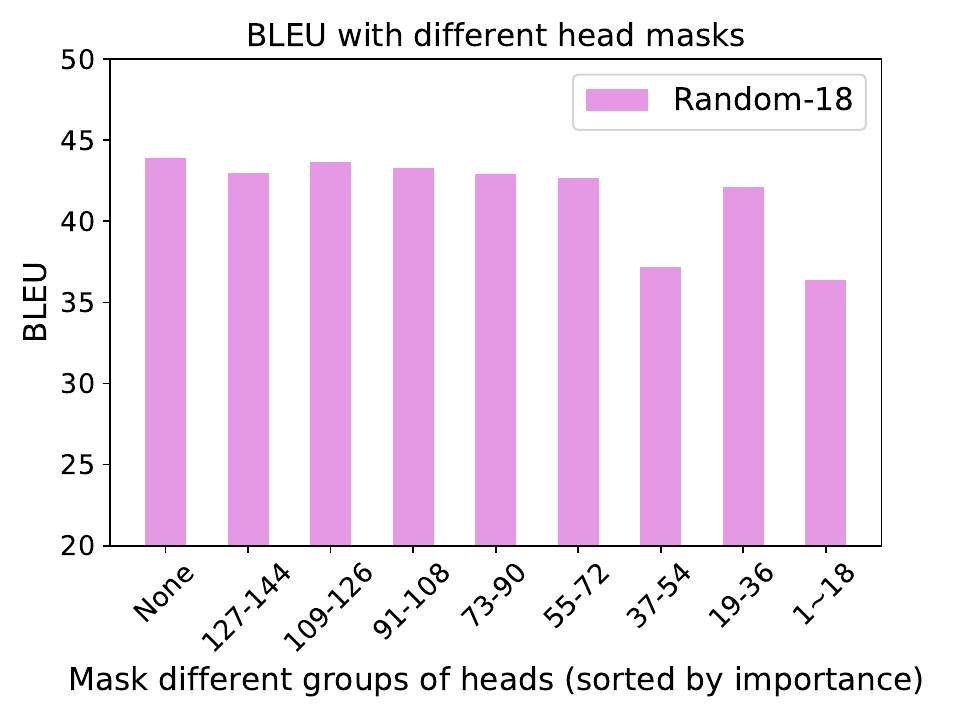}
        \end{minipage}
    }
    \subfigure[Impt-18]{
        \begin{minipage}{.31\textwidth}
            \centering
            \includegraphics[scale = 0.3]{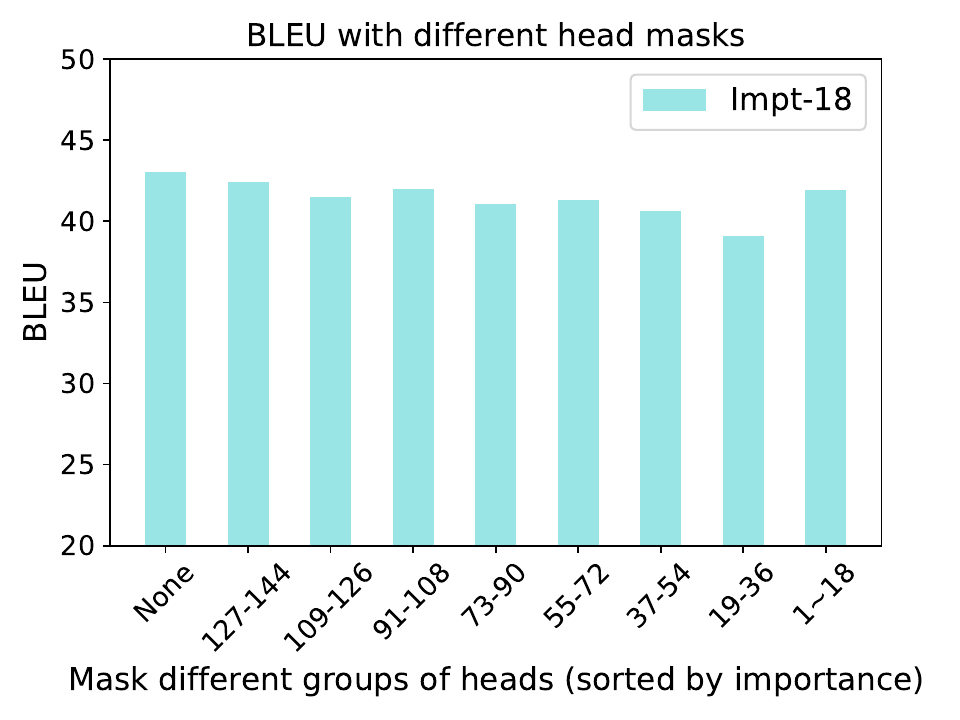}
        \end{minipage}
    }
    \caption{Our methods significantly maintain the performance even if the important heads are masked.}
    \label{fig_maskhead_all}
\end{figure*}

\begin{figure*}[h]
    \centering
    \subfigure[Mask 18 heads in training]{
        \begin{minipage}{.31\textwidth}
            \centering
            \includegraphics[scale = 0.3]{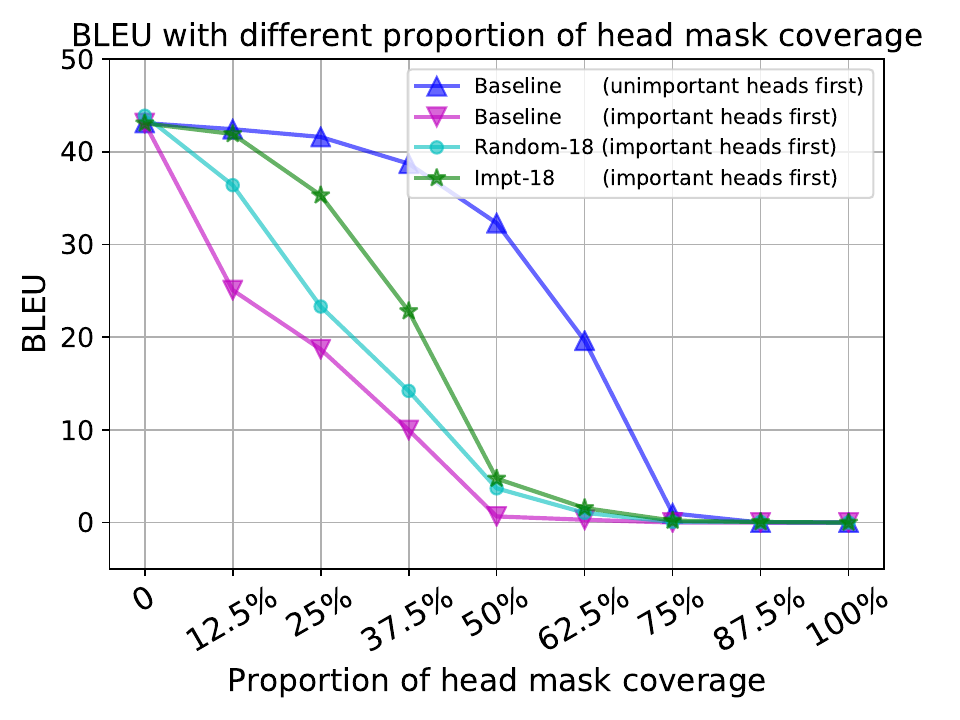}
        \end{minipage}
    }
    \subfigure[Mask 36 heads in training]{
        \begin{minipage}{.31\textwidth}
            \centering
            \includegraphics[scale = 0.3]{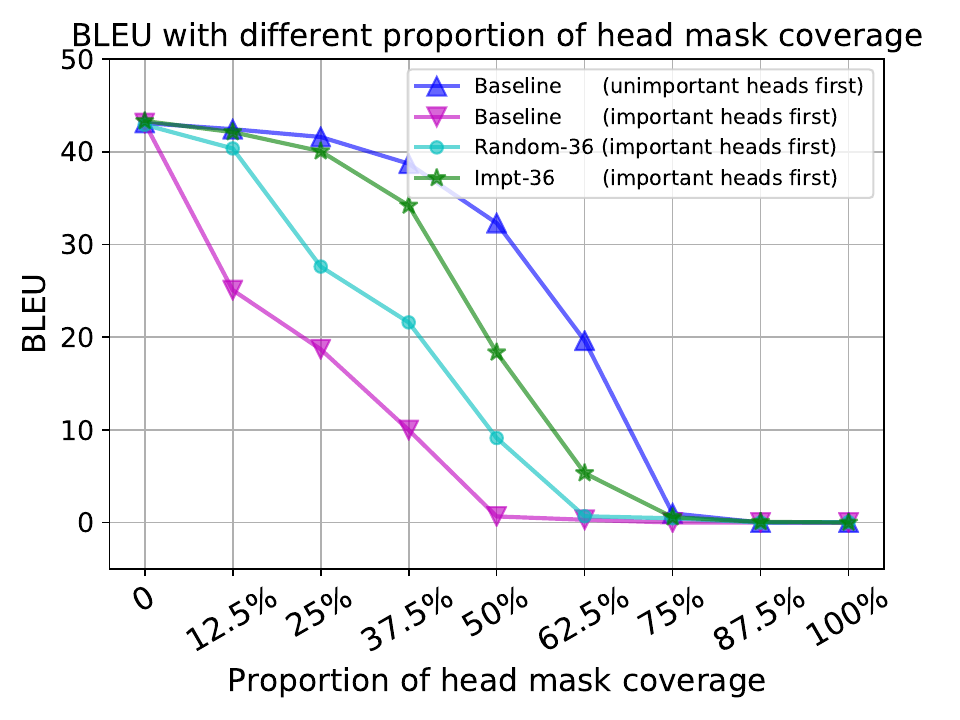}
        \end{minipage}
    }
    \subfigure[Mask 54 heads in training]{
        \begin{minipage}{.31\textwidth}
            \centering
            \includegraphics[scale = 0.3]{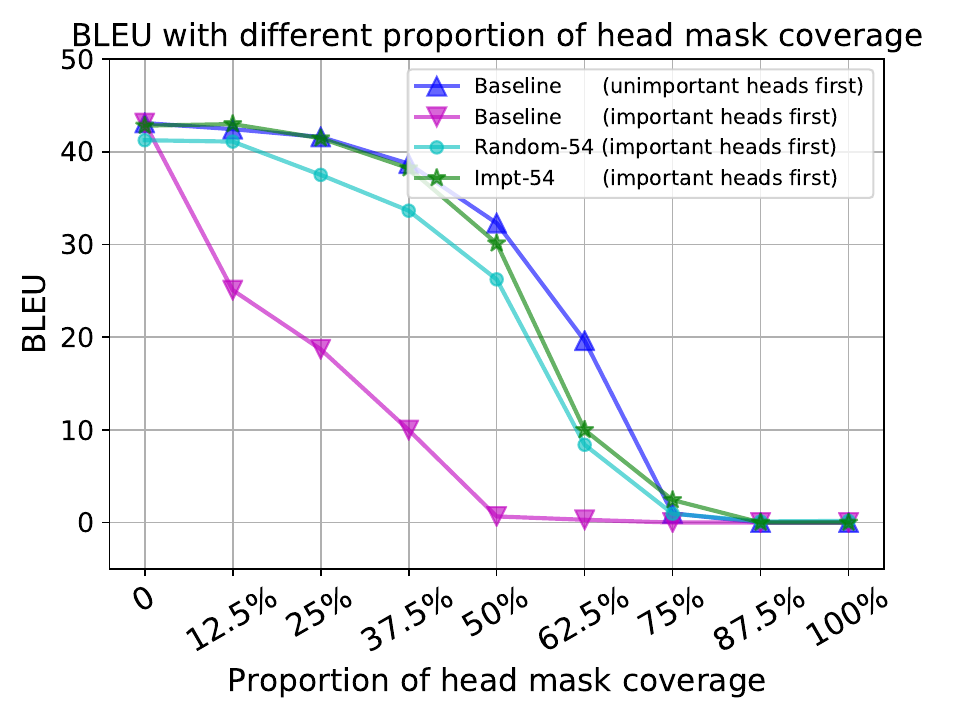}
        \end{minipage}
    }
    \caption{As the number of masked heads grows, the drop curves starting with important heads are moving up.}
    \label{fig_reduce_all}
\end{figure*}

\paragraph{Comparison} We compare the baseline with masking randomly (Random-N) and masking important ones (Impt-N), where N is the mask number. In this paper, we mainly employ $N=18 (12.5\%)$.


\subsection{Results} 
As is shown in Table~\ref{tab_trans_zh2en},\ref{tab_trans_x2en},\ref{tab_trans_en2x}, except for Vi-En experiments, Impt-18 yields enhancement over all language directions and reach the best result on the experiment of Ro $\rightarrow$ En. And Random-18 obtains steady improvements over all pairs and is obviously better than Impt-18. It seems the aggressive masking strategy at important heads can be too harsh and reversely restrict the model. And the random method is more expert in building a rational training pattern. In conclusion, reducing the unbalanced training among attention heads can effectively improve the translation quality.

\begin{table}[h]\footnotesize
    \centering
    \begin{tabular}{lccccc}
        \hline
        Test sets & MT04 & MT05 & MT06 \\ \hline
        Baseline & 46.62 & 43.46 & 43.09 \\
        Impt-18 & 46.94 (+0.28) & 44.19 (+0.73) & 43.16 (+0.07) \\
        Random-18 & \textbf{47.04 (+0.42)} & \textbf{44.33 (+0.87)} & \textbf{43.88 (+0.79)} \\ \hline
    \end{tabular}
    \caption{Results on Experiments of Zh $\rightarrow$ En}
    \label{tab_trans_zh2en}
\end{table}

\begin{table}[h]\footnotesize
    \centering
    \begin{tabular}{lccc}
        \hline
        Directions & Ro $\rightarrow$ En & Vi $\rightarrow$ En  & Tr $\rightarrow$ En \\ \hline
        Baseline & 32.17 & 26.49 & 17.29 \\
        Impt-18 & \textbf{32.95 (+0.78)} & 26.36 (-0.13) & 17.48 (+0.19) \\
        Random-18 & 32.85 (+0.68) & \textbf{26.85 (+0.36)} & \textbf{17.56 (+0.27)} \\ \hline
    \end{tabular}
    \caption{Results on Experiments of Ro/Vi/Tr $\rightarrow$ En}
    \label{tab_trans_x2en}
\end{table}

\begin{table}[h]\footnotesize
    \centering
    \begin{tabular}{lccc}
        \hline
        Directions & En $\rightarrow$ Ro & En $\rightarrow$ Vi & En $\rightarrow$ Tr \\ \hline
        Baseline & 31.98 & 28.07 & 15.74 \\
        Impt-18 & 32.47 (+0.49) & 28.06 (-0.01) & 16.10 (+0.36) \\
        Random-18 & \textbf{32.64 (+0.66)} & \textbf{28.46 (+0.39)} & \textbf{16.16 (+0.42)} \\ \hline
    \end{tabular}
    \caption{Results on Experiments of En $\rightarrow$ Ro/Vi/Tr}
    \label{tab_trans_en2x}
\end{table}

\subsection{Statistical Analysis} 

\subsubsection{Flatter Distribution} 

To evaluate the adjusted training of heads, we check the distribution of head importance. As is shown in Figure~\ref{fig_maskhead_dis}, our methods make the importance distribution flatter. And the overall variance and mean are also calculated, as is shown in Table~\ref{tab_maskhead_var},\ref{tab_maskhead_mean}. Compared with Baseline, Impt-18 and Random-18 significantly reduce the variance of attention heads, achieving the goal of more equal training. And the mean also decreases, which proves the decline of dependence on every individual head. More specifically, Impt-18 can better resolve the imbalance, for it well prevent the emergence of ``super'' heads.


\begin{table}[h]\footnotesize
    \centering
    \begin{tabular}{lcccc}
        \hline
        Directions & Zh2En & Ro2En & Vi2En & Tr2En\\ \hline
        Baseline & 77.28 & 552.93 & 100.73 & 1767.70 \\
        Random-18 & 33.21 & 255.98 & 48.28 & 900.70  \\
        Impt-18 & 9.13 & 72.73 & 14.13 & 188.87  \\ \hline
    \end{tabular}
    \captionsetup{font={footnotesize}}
    \caption{Our methods greatly reduce the \textbf{Variance} of the head importance, illustrating the improved equality of heads.}
    \label{tab_maskhead_var}
\end{table}


\begin{table}[h]\footnotesize
    \centering
    \begin{tabular}{lcccc}
        \hline
        Directions & Zh2En & Ro2En & Vi2En & Tr2En\\ \hline
        Baseline & 27.15 & 47.18 & 17.96 & 83.79 \\ 
        Random-18 & 19.62 & 39.96 & 14.86 & 74.05  \\
        Impt-18 & 18.95 & 37.30 & 18.96 & 85.12  \\ \hline
    \end{tabular}
    \captionsetup{font={footnotesize}}
    \caption{Our methods reduce the \textbf{Mean} of the head importance, illustrating the lessened dependence on each head.}
    \label{tab_maskhead_mean}
\end{table}

\subsubsection{Weaker Dependence}

We repeat the experiments of masking different groups of heads. As is shown in Figure~\ref{fig_maskhead_all}, the translation quality is still maintained even if important heads are masked, proving the dependence on them has decreased. And Impt-18 performs more steadily since it is accustomed to such situations.

\subsubsection{More Robust Models}

We also repeat the experiments of masking all heads, as is shown in Figure~\ref{fig_reduce_all}. The two middle lines originally lie in the same place as the bottom one. As the number of masked heads in training (N) grows, they gradually move up and approach the top line where unimportant heads are masked first. It shows our methods make the model rely less on the important heads and become more robust.

\section{Related Works}
 
Recently, many analytical works about multi-head attention come out~\cite{Raganato2018AnAO,tang2018analysis,Voita2019AnalyzingMS,Michel2019AreSH,sun2020generating,Behnke2020LosingHI}. And for the inequality of the networks, some studies focus on the model level~\cite{frankle2019lottery,sun2021multilingual}, layer level~\cite{zhang2019improving}, and neuron level~\cite{bau2019identifying}. For the mask algorithm, there are also works on the layer level~\cite{fan2020reducing}, word level~\cite{provilkov2019bpe}, and neuron level~\cite{Srivastava2014DropoutAS}. Different from them, we mainly study the attention level and conduct a statistical analysis.

\section{Conclusion}

In this paper, we empirically validate the inequality of attention heads in Transformer and come up with an assumption of imbalanced training. Correspondingly, we propose a specific method in two ways to resolve the issue. Experiments show the improvements on multiple language pairs. And detailed analysis shows the alleviation of the problem and the effectiveness of our techniques.

\section{Acknowledgements}
We would like to thank the anonymous reviewers for their insightful comments. Shujian Huang is the corresponding author. This work is supported by National Science Foundation of China (No. 6217020152).

\bibliography{acl}
\bibliographystyle{acl_natbib}




\end{document}